\documentstyle[12pt]{article}
\begin{document}

\title{On Using Selectional Restriction in Language Models for Speech
Recognition}
\author{\\Joerg P. Ueberla\\ \\ \\
CMPT TR 94-03\\
School of Computing Science\thanks{This is a revised version of the original
technical report,
prepared for the cmp-lg server.
The work was supported by NSERC operating grant OGP0041910. },
Simon Fraser University,\\
Burnaby, B.C.,
V5A 1S6,
CANADA\\
email: ueberla@cs.sfu.ca
}

\maketitle

\begin{abstract}

In this paper, we investigate the use of selectional restriction -- the
constraints a
predicate imposes on its arguments -- in a language model for speech
recognition. We
use an un-tagged corpus, followed by a public domain tagger and a very simple
finite
state machine to obtain verb-object pairs from unrestricted English text. We
then
measure the impact the knowledge of the verb has on the prediction of the
direct object
in terms of the perplexity of a cluster-based language model. The results show
that even
though a clustered bigram is more useful than a verb-object model, the
combination of the
two leads to an improvement over the clustered bigram model.
\pagebreak
\end{abstract}

\section{Introduction}

The constraint a predicate imposes on its arguments is called selectional
restriction.
For example, in the sentence fragment ``She eats x'', the verb ``eat'' imposes
a constraint
on the direct object ``x'': ``x'' should be something that is usually being
eaten. This
phenomenon has received considerable attention in the linguistic community and
it was
recently explained in statistical terms in \cite{Res93}. Because it restricts
subsequent nouns and because nouns account for a large part of the perplexity
(see \cite{Ueb94}), it seems natural to try to
use selectional restriction in a language model for speech recognition. In this
paper,
we report on our work in progress on this topic. We begin by  presenting in
section
\ref{data}
the process we use to obtain training and testing data from unrestricted
English text.
The data constitutes the input to a clustering algorithm and a language model,
both of which
are described in section \ref{lm}. In section \ref{results}, we present the
results we have obtained so far, followed by conclusions in section
\ref{conclusions}.

\section{Training and Testing Data}
\label{data}

\begin{figure}
\begin{center}
\setlength{\unitlength}{0.012500in}%
\begingroup\makeatletter\ifx\SetFigFont\undefined
\def\x#1#2#3#4#5#6#7\relax{\def\x{#1#2#3#4#5#6}}%
\expandafter\x\fmtname xxxxxx\relax \def\y{splain}%
\ifx\x\y   
\gdef\SetFigFont#1#2#3{%
  \ifnum #1<17\tiny\else \ifnum #1<20\small\else
  \ifnum #1<24\normalsize\else \ifnum #1<29\large\else
  \ifnum #1<34\Large\else \ifnum #1<41\LARGE\else
     \huge\fi\fi\fi\fi\fi\fi
  \csname #3\endcsname}%
\else
\gdef\SetFigFont#1#2#3{\begingroup
  \count@#1\relax \ifnum 25<\count@\count@25\fi
  \def\x{\endgroup\@setsize\SetFigFont{#2pt}}%
  \expandafter\x
    \csname \romannumeral\the\count@ pt\expandafter\endcsname
    \csname @\romannumeral\the\count@ pt\endcsname
  \csname #3\endcsname}%
\fi
\fi\endgroup
\begin{picture}(320,292)(40,505)
\thinlines
\put(255,715){\vector(-1, 0){120}}
\put(220,600){\vector( 1, 1){ 82.500}}
\put(324,676){\vector(-1,-1){ 83}}
\put(115,680){\vector( 1,-1){ 84}}
\put(174,595){\vector(-1, 1){ 84.500}}
\put(300,725){\circle{36}}
\put( 95,725){\circle{36}}
\put(210,565){\circle{36}}
\put( 75,725){\line(-1, 0){ 35}}
\put( 40,725){\line( 0, 1){ 35}}
\put( 40,760){\line( 1, 0){ 55}}
\put( 95,760){\vector( 0,-1){ 15}}
\put(320,725){\line( 1, 0){ 40}}
\put(360,725){\line( 0, 1){ 35}}
\put(360,760){\line(-1, 0){ 60}}
\put(300,760){\vector( 0,-1){ 15}}
\put(230,565){\line( 1, 0){ 40}}
\put(270,565){\line( 0,-1){ 30}}
\put(270,535){\line(-1, 0){ 55}}
\put(215,535){\vector( 0, 1){ 15}}
\put(135,740){\vector( 1, 0){120}}
\put( 90,715){\makebox(0,0)[lb]{\smash{\SetFigFont{20}{24.0}{rm}1}}}
\put(240,505){\makebox(0,0)[lb]{\smash{\SetFigFont{20}{24.0}{rm}PP}}}
\put(295,715){\makebox(0,0)[lb]{\smash{\SetFigFont{20}{24.0}{rm}2}}}
\put(205,555){\makebox(0,0)[lb]{\smash{\SetFigFont{20}{24.0}{rm}3}}}
\put(185,690){\makebox(0,0)[lb]{\smash{\SetFigFont{20}{24.0}{rm}NC}}}
\put(165,645){\makebox(0,0)[lb]{\smash{\SetFigFont{20}{24.0}{rm}PP}}}
\put(235,650){\makebox(0,0)[lb]{\smash{\SetFigFont{20}{24.0}{rm}V}}}
\put(310,625){\makebox(0,0)[lb]{\smash{\SetFigFont{20}{24.0}{rm}PP}}}
\put(185,765){\makebox(0,0)[lb]{\smash{\SetFigFont{20}{24.0}{rm}V}}}
\put( 90,610){\makebox(0,0)[lb]{\smash{\SetFigFont{20}{24.0}{rm}NC}}}
\put( 40,775){\makebox(0,0)[lb]{\smash{\SetFigFont{20}{24.0}{rm}NC}}}
\put(355,770){\makebox(0,0)[lb]{\smash{\SetFigFont{20}{24.0}{rm}V}}}
\end{picture}
\caption{Finite State Automaton}
\label{fig:fsa}
\end{center}
\end{figure}
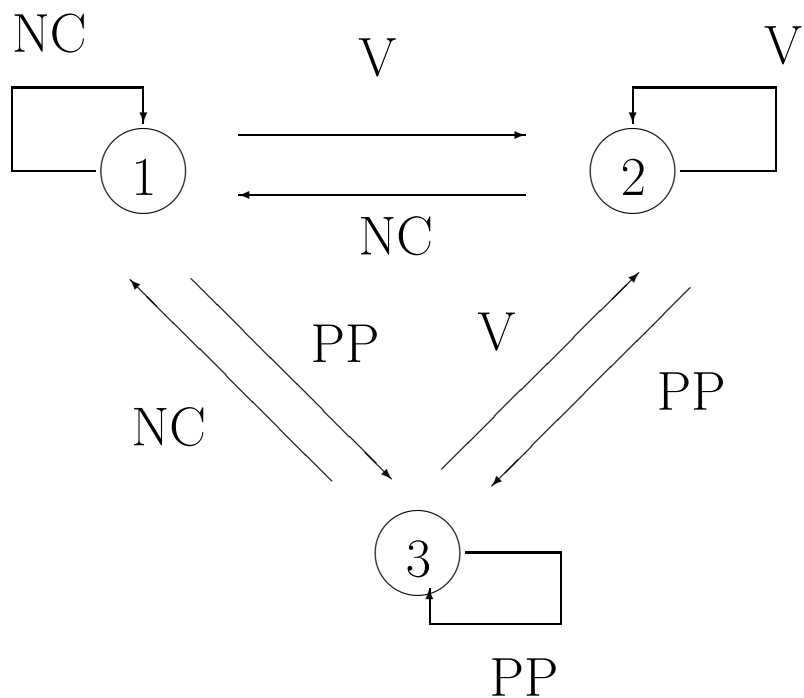

In order to use  selectional restriction in a language model for speech
recognition, we
have to be able to identify the predicate and its argument in naturally
occurring,
unrestricted English text in an efficient manner. Since the parsing of
unrestricted text
is a yet unsolved, complicated problem by itself, we do not attempt to use a
sophisticated
parser. Instead, we use the un-tagged version of the Wall Street
Journal Corpus (distributed by ACL/DCI), Xerox's public domain tagger
(described in \cite{Cut92}) and a very simple deterministic
finite-state automaton to identify verbs and their direct objects.
The resulting data is certainly very noisy, but, as opposed to more
accurate data obtained from a sophisticated parser,  it would be feasible to
use
this method in a speech recogniser. The finite-state automaton we use only
has three states and it is shown in Figure \ref{fig:fsa}.
The circles correspond
to states and the arcs to transitions.
The input to the automaton
consists of a sequence of words with associated tags. The words and tags
\footnote{``v'' corresponds to most forms of most verbs,
``b'' corresponds to forms of ``to be'',
``h'' corresponds to forms of ``to have'',
``cs'' contains words like ``although'', ``since'' and
``w'' contains words like ``who'', ``when''. The tagset we use is the
one provided by the tagger and for more information please refer to
\cite{Cut92}.
}
are classified into
three different events:
\begin{itemize}
\item V : the word is a verb (iff its tag starts with ``v'', ``b'' or ``h'')
\item PP: the word is a preposition (iff its tag is ``in'')
\item NC: the word starts a new clause (iff its tag is
``.'',``:'',``;'',``!'',``?'',``cs'' or
begins with ``w'')
\end{itemize}
All other words and tags do not lead to a change in the state of the automaton.
Intuitively, state 1 corresponds to the beginning of a new clause without
having seen its verb,
state 2 to a clause after having seen its verb
and state 3 to a prepositional phrase.
Given this automaton, we then consider  each occurrence
of a noun (e.g. tag ``nn'')  to be
\begin{itemize}
\item a direct object if we are currently in state 2; the corresponding verb is
considered to be the one that caused the last transition into state 2;
\item part of  a prepositional phrase if we are currently in state 3; the
corresponding
preposition is considered to be the one that caused the last transition into
state 3;
\item unconstrained by a predicate or preposition (for example a subject) if we
are in state 1.
\end{itemize}
The automaton then outputs a sequence of verb-object pairs, which constitute
our training and testing data.
Examples of the output of the automaton  are shown in Table \ref{tab:data}.
Entries that would not normally be considered verb-object pairs are marked with
``*''.
\begin{table}
\centering
\begin{tabular}{|ccc|} \hline
Verb & Object &\\ \hline
join & board & \\
is & chairman & * \\
named & director &  \\
make & cigarette & \\
make & filters  &\\
caused & percentage & \\
enters & lungs & \\
causing & symptoms & \\
show & decades & *\\
show & researchers & \\
Loews & Corp. & *\\
makes & cigarettes & \\
stopped & crocidolite & \\
bring & attention & \\
This & old & * \\
having & properties & \\
is & asbestos & * \\
studied & workers  & \\ \hline
\end{tabular}
\caption{Examples of the noisy verb-object data}
\label{tab:data}
\end{table}
The data is very noisy because errors can be
introduced both by the tagger (which makes its decisions based on a small
window of words) and
by the overly simplistic finite state automaton (for example, not all
occurrences of ``cs''
and ``w'' constitute a new clause). Given this data, the goal of our language
model is to
predict the direct objects and we will measure the influence
the knowledge of the preceding verb has on this prediction in terms of
perplexity.

\section{The Language Model}
\label{lm}

We can formulate the task of the language model as the
prediction of  the value of some variable $Y$ (e.g. the
next word) based on some knowledge about the past encoded by
variables $X_{1}, ..., X_{n}$.
In our case, $Y$ is the next direct object and the only knowledge we have about
the past is
the identity of the previous verb encoded by the variable $X$. The most
straight forward way
would be to directly estimate the conditional probability distribution
$p(Y=y_{l}|X=x_{k})$.
However, because
of sparse training data, it is often difficult to estimate this distribution
directly.
Class based models, that group elements $x_{k} \in {\cal X}$ into classes
$g_{x}=G_{1}(x_{k})$
and elements $y_{l} \in {\cal Y}$
into classes $g_{y}=G_{2}(y_{l})$ can be used to alleviate this problem. The
conditional probability distribution
is then calculated as
\begin{equation}
 p_{G}(y_{l}|x_{k})=p(G_{2}(y_{l})|G_{1}(x_{k})) * p(y_{l}|G_{2}(y_{l}))
\label{eq:one},
\end{equation} which generally requires less
training data
\footnote{As an example, if ${\cal X}$ and  ${\cal Y}$ have 10,000 elements
each, and if we
use $200$ classes for $G_{1}$ and for $G_{2}$, then the original model has to
estimate
$10,000*10,000=1*10^{8}$ probabilities whereas the class based model only needs
to
estimate $200*200 + 200*10,000=2.04*10^{6}$.}.
In the following, let $(X[i], Y[i]), 1 \leq i \leq N$ denote the values of $X$
and $Y$ at the
$i^{th}$ data point and let $(G_{1}[i], G_{2}[i])$ denote the corresponding
classes.
How can we obtain the classification functions $G_{1}$ and $G_{2}$
automatically, given only
a set of $N$ data points $(X[i], Y[i])$ ?
In the following (sections \ref{ml}, \ref{lo}, \ref{algo}), we describe a
method which is almost identical to
the one presented in  \cite{Kne93}.
The only  difference is that
in \cite{Kne93},  the elements of variables $X$ and $Y$ are identical (the data
consists of bigrams),
thus requiring only one clustering function $G$
\footnote{It is possible that using two clustering functions would be
beneficial even if the two
variables have the same elements.}. In our case, however, the
variables $X$ and $Y$ contain different elements (verbs and objects
respectively), and we thus
produce two clustering functions $G_{1}$ and $G_{2}$.

\subsection{Maximum-Likelihood Criterion}
\label{ml}

In order to automatically find classification functions $G_{1}$ and $G_{2}$,
which -- as a shorthand --
 we will also denote as $G$, we first convert the classification problem into
an optimisation problem.
Suppose the function $F(G)$ indicates how good the classification $G$ (composed
of $G_{1}$ and $G_{2}$) is. We can then
reformulate the classification problem as finding the classification $G$ that
maximises F:
\begin{equation} G = argmax_{G^{'} \in {\cal G}} F(G^{'}),
\end{equation}
where ${\cal G}$ contains the set of possible classifications which are at our
disposal.

What is a suitable function $F$, also called optimisation criterion? Given a
classification function
$G$, we can estimate the probabilities $p_{G}(y_{l}|x_{k})$ of equation
\ref{eq:one} using
the maximum likelihood estimator, e.g. relative frequencies:
\begin{eqnarray}
p_{G}(y_{l}|x_{k}) & = & p(G_{2}(y_{l})|G_{1}(x_{k})) * p(y_{l}|G_{2}(y_{l}))\\
 & = & \frac{N(g_{x}, g_{y})}{N(g_{x})} * \frac{N(g_{y}, y)}{N(g_{y})},
\end{eqnarray}
where $g_{x}=G_{1}(x_{k})$, $g_{y}=G_{2}(y_{l})$ and $N(x)$ denotes the number
of times $x$ occurs in the data.
Given these probability estimates $p_{G}(y_{l}|x_{k})$, the likelihood $F_{ML}$
of the training data, e.g. the probability of the training data being generated
by our probability
estimates $p_{G}(y_{l}|x_{k})$, measures how well the training data is
represented by the
estimates and can be used as optimisation criterion (\cite{Jel90}).

In the following, we will derive an optimisation function $F_{ML}$ in terms
of frequency counts observed in the training data.
The likelihood of the
training data $F_{ML}$ is simply
\begin{eqnarray}
F_{ML} & = & \prod_{i=1}^{N} p_{G}(Y[i]|X[i])\\
 & = &  \prod_{i=1}^{N} \frac{N(G_{1}[i], G_{2}[i])}{N(G_{1}[i])} *
\frac{N(G_{2}[i], Y[i])}{N(G_{2}[i])}.
\end{eqnarray}
Assuming that the classification is unique, e.g. that $G_{1}$ and $G_{2}$ are
functions,
we have $N(G_{2}[i], Y[i])=N(Y[i])$
(because $Y[i]$ always occurs with the same class $G_{2}[i]$). Since we try to
optimise $F_{ML}$ with respect to $G$, we
can remove any term that does not depend on $G$, because it will not influence
our optimisation. It is
thus equivalent to optimise
\begin{eqnarray}
F^{'}_{ML} & = & \prod_{i=1}^{N} f(X[i], Y[i])\\
 & = & \prod_{i=1}^{N} \frac{N(G_{1}[i], G_{2}[i])}{N(G_{1}[i])} *
\frac{1}{N(G_{2}[i])} .
\end{eqnarray}
If, for two pairs $(X[i], Y[i])$ and $(X[j], Y[j])$, we have
$G_{1}(X[i])=G_{1}(X[j])$ and
$G_{2}(Y[i])=G_{2}(Y[j])$, then we know that $f(X[i], Y[i])=f(X[j], Y[j])$.
We can thus regroup identical terms to obtain
\begin{equation}
 F^{'}_{ML}= \prod_{g_{x}, g_{y}} [ \frac{N(g_{x}, g_{y})}{N(g_{x})} *
\frac{1}{N(g_{y})} ] ^{N(g_{x}, g_{y})},
\end{equation}
where the product is over all possible pairs $(g_{x}, g_{y})$.
Because $N(g_{x})$ does not depend on $g_{y}$ and $N(g_{y})$ does not depend on
$g_{x}$, we can
simplify this again to
\begin{equation}
F^{'}_{ML}= \prod_{g_{x}, g_{y}} N(g_{x}, g_{y})^{N(g_{x}, g_{y})}
\prod_{g_{x}} \frac{1}{N(g_{x})}^{N(g_{x})}
\prod_{g_{y}} \frac{1}{N(g_{y})}^{N(g_{y})}.
\end{equation}
Taking the logarithm, we obtain the equivalent optimisation criterion
\begin{eqnarray}
 F^{''}_{ML} & = &  \sum_{g_{x}, g_{y}}  N(g_{x}, g_{y}) * log( N(g_{x}, g_{y}
)) -
\sum_{g_{x}} N(g_{x}) * log ( N(g_{x}))\\
 &  - &  \sum_{g_{y}} N(g_{y}) * log ( N(g_{y})) \nonumber .
\end{eqnarray}

$F_{ML}$ is the maximum likelihood optimisation criterion and we could use it
to find good classifications
$G$. However, the problem with this maximum likelihood criterion is that we
first estimate the probabilities
$p_{G}(y_{l}|x_{k})$ on the training data $T$ and then, given
$p_{G}(y_{l}|x_{k})$, we evaluate
the classification $G$ on $T$. In other words, both the classification $G$ and
the estimator
$p_{G}(y_{l}|x_{k})$ are trained on the same data. Thus, there will not be any
unseen event, a fact
that overestimates the power for generalisation of the class based model. In
order to avoid this, we will
in section \ref{lo}
incorporate a cross-validation technique directly into the optimisation
criterion.

\subsection{Leaving-One-Out Criterion}
\label{lo}

The basic principle of cross-validation is to split the training data $T$ into
a ``retained'' part
$T_{R}$ and a ``held-out'' part $T_{H}$. We can then use $T_{R}$ to estimate
the probabilities
$p_{G}(y_{l}|x_{k})$ for a given classification $G$,  and $T_{H}$  to evaluate
how well the
classification $G$ performs. The so-called
leaving-one-out technique is a special case of cross-validation
(\cite[pp.75]{Dud73}).
It divides the data into $N-1$ samples as ``retained'' part and only one sample
as ``held-out'' part. This is repeated $N-1$ times, so that each sample is
once in the ``held-out'' part. The advantage of this approach is that all
samples
are used in the ``retained'' and in the ``held-out'' part, thus making very
efficient
use of the existing data. In other words, our  ``held-out'' part $T_{H}$ to
evaluate
a classification $G$ is the entire set of data points; but when we calculate
the
probability of the $i^{th}$ data point, we assume that our probability
distribution
$p_{G}(y_{l}|x_{k})$ was estimated on all the data expect point $i$.

Let $T_{i}$ denote the data without the pair $(X[i], Y[i])$ and
$p_{G,T_{i}}(y_{l}|x_{k})$ the probability estimates based on a given
classification $G$ and
training corpus $T_{i}$. Given a particular $T_{i}$, the probability of the
``held-out''
part $(X[i], Y[i])$ is $p_{G,T_{i}}(y_{l}|x_{k})$. The probability of the
complete corpus,
where each pair is in turn considered the ``held-out'' part is the
leaving-one-out likelihood $L_{LO}$
\begin{equation}
 L_{LO}=\prod_{i=1}^{N} p_{G,T_{i}}(Y[i]|X[i]).
\label{eq:two}
\end{equation}
In the following, we will derive an optimisation function $F_{LO}$ by
specifying how
$p_{G,T_{i}}(Y[i]|X[i])$ is estimated from frequency counts.
 First we rewrite $p_{G,T_{i}}(Y[i]|X[i])$  as usual (see equation
\ref{eq:one}):
\begin{eqnarray}
 p_{G,T_{i}}(y_{l}|x_{k}) & = & P_{G,T_{i}}(G_{2}(y_{l})|G_{1}(x_{k}))*P_{G,
T_{i}}(y_{l}|G_{2}(y_{l}))\\
 & = & \frac{p_{G,T_{i}}(g_{x}, g_{y})}{p_{G,T_{i}}(g_{x})} *
\frac{p_{G,T_{i}}(g_{y}, y_{l})}{p_{G,T_{i}}(g_{y})},
\label{eq:estim}
\end{eqnarray}
where $g_{x}=G_{1}(x_{k})$ and $g_{y}=G_{2}(y_{l})$.
Now we will specify how we estimate each term in equation \ref{eq:estim}.

As we saw before, $p_{G,T_{i}}(g_{y}, y_{l})=p_{G,T_{i}}(y_{l})$ (if the
classification $G_{2}$
is a function) and since $p_{T_{i}}(y_{l})$ is actually independent of $G$, we
can drop it
out of the maximization and thus need not specify an estimate for it.

As we will see later, we can guarantee that every class
$g_{x}$ and $g_{y}$ has been seen at least once in the
``retained'' part and we can thus use relative counts as estimates for class
uni-grams:
\begin{eqnarray}
p_{G,T_{i}}(g_{x}) & = & \frac{N_{T_{i}}(g_{x})}{N_{T_{i}}} \label{eq:three}\\
p_{G,T_{i}}(g_{y}) & = & \frac{N_{T_{i}}(g_{y})}{N_{T_{i}}} \label{eq:four} .
\end{eqnarray}

In the case of the class bi-gram, we might have to predict unseen events
\footnote{If $(X[i], Y[i])$ occurs only once in the complete corpus, then
$p_{G,T_{i}}(Y[i]|X[i])$ will have to be calculated based on the corpus
$T_{i}$, which does not
contain any occurrences of $(X[i], Y[i])$.}.
We therefore use the absolute discounting
method (\cite{Ney93}), where the counts are reduced by a constant value $b < 1$
and where the
gained probability mass is redistributed over unseen events. Let
$n_{0,T_{i}}$ be the number of unseen pairs $(g_{x}, g_{y})$ and $n_{+,T_{i}}$
the number of
seen pairs $(g_{x}, g_{y})$, leading to the following smoothed estimate
\begin{eqnarray}
\lefteqn{ p_{G,T_{i}}(g_{x}, g_{y})} \nonumber \\
 & = & \left\{ \begin{array}{ll}
\frac{N_{T_{i}}(g_{x}, g_{y}) - b}{N_{T_{i}}} & \mbox{if $N_{T_{i}}(g_{x},
g_{y})>0$}\\
\frac{n_{+, T_{i}}*b}{n_{0,T_{i}}*N_{T_{i}}} & \mbox{if $N_{T_{i}}(g_{x},
g_{y})=0$}\\
\end{array}
\right.
\end{eqnarray}
Ideally, we would make $b$ depend on the classification, e.g. use
$b=\frac{n_{1}}{n_{1}+2*n_{2}}$, where
$n_{1}$ and $n_{2}$ depend on $G$. However, due to computational reasons, we
use, as suggested in
\cite{Kne93}, the empirically determined constant value $b=0.75$ during
clustering.
The probability distribution $p_{G,T_{i}}(g_{x}, g_{y})$ will always be
evaluated on the
``held-out'' part $(X[i], Y[i])$ and with $g_{x,i}=G_{1}[i]$ and
$g_{y,i}=G_{2}[i]$ we obtain
\begin{eqnarray}
\lefteqn{ p_{G,T_{i}}(g_{x,i}, g_{y,i})} \nonumber \\
 & = & \left\{ \begin{array}{ll}
\frac{N_{T_{i}}(g_{x,i}, g_{y,i}) - b}{N_{T_{i}}} & \mbox{if
$N_{T_{i}}(g_{x,i}, g_{y,i})>0$}\\
\frac{n_{+, T_{i}}*b}{n_{0,T_{i}}*N_{T_{i}}} & \mbox{if $N_{T_{i}}(g_{x,i},
g_{y,i})=0$}
\label{eq:five}
\end{array}
\right.
\end{eqnarray}

In order to facilitate future regrouping of terms, we can now express the
counts
$N_{T_{i}}, N_{T_{i}}(g_{x})$ etc.
in terms of the counts of
the complete corpus $T$ as follows:
\begin{eqnarray}
N_{T_{i}} & = & N_{T} - 1 \label{eq:six} \\
N_{T_{i}}(g_{x}) & = & N_{T}(g_{x}) - 1 \\
N_{T_{i}}(g_{y}) & = & N_{T}(g_{y}) - 1 \\
N_{T_{i}}(g_{x,i}, g_{y,i}) & = &  N_{T}(g_{x,i}, g_{y,i})-1\\N_{T_{i}} & = &
N_{T} - 1 \\
n_{+, T_{i}} & = & \left\{
\begin{array}{ll}
n_{+, T} & \mbox{if $N_{T}(g_{x,i}, g_{y,i})>1$}\\
n_{+, T} - 1  & \mbox{if $N_{T}(g_{x,i}, g_{y,i})=1$}\\
\end{array} \right. \\
n_{0,T_{i}} & = &  \left\{
\begin{array}{ll}
n_{0,T} & \mbox{if $N_{T}(g_{x,i}, g_{y,i})>1$}\\
n_{0, T} - 1  & \mbox{if $N_{T}(g_{x,i}, g_{y,i})=1$}
\label{eq:sixb}
\end{array} \right.
\end{eqnarray}
All we have left to do now is to substitute all the expressions back into
equation \ref{eq:two}.
After dropping $p_{G,T_{i}}(y_{l})$ because it is independent of $G$ we get
\begin{eqnarray}
F'_{LO} & = & \prod_{i=1}^{N} \frac{p_{G,T_{i}}(g_{x,i},
g_{y,i})}{p_{G,T_{i}}(g_{x,i})} * \frac{1}{p_{G,T_{i}}(g_{y,i})}\\
& = & \prod_{g_{x}, g_{y}} ( p_{G,T_{i}}(g_{x}, g_{y}))^{N(g_{x}, g_{y})} *
\prod_{g_{x}} (\frac{1}{p_{G,T_{i}}(g_{x})})^{N(g_{x})} *
\prod_{g_{y}} (\frac{1}{p_{G,T_{i}}(g_{y})})^{N(g_{y})} .
\end{eqnarray}
We can now substitute equations \ref{eq:three}, \ref{eq:four} and
\ref{eq:five}, using
the counts of the whole corpus of equations \ref{eq:six} to \ref{eq:sixb} .
After having dropped
terms independent of $G$, we obtain
\begin{eqnarray}
F''_{LO} & = & \prod_{g_{x}, g_{y} : N(g_{x},g_{y})  > 1 } (N_{T}(g_{x}, g_{y})
-1 -b )^{N_{T}(g_{x}, g_{y})} *
\left( \frac{(n_{+,T}-1)*b}{(n_{0,T}+1)} \right)^{n_{1,T}}\\
 &  * & \prod_{g_{x}} \left( \frac{1}{(N_{T}(g_{x}-1))} \right)^{N_{T}(g_{x})}
* \prod_{g_{y}} \left( \frac{1}{(N_{T}(g_{y}-1))} \right)^{N_{T}(g_{y})}
\nonumber ,
\end{eqnarray}
where $n_{1,T}$ is the number of pairs $(g_{x}, g_{y})$ seen exactly once in
$T$
(e.g. the number of pairs that will be unseen when used as ``held-out'' part).
Taking the logarithm, we obtain the final optimisation criterion $F'''_{LO}$
\begin{eqnarray}
F'''_{LO} & = & \sum_{g_{x}, g_{y}:N_{T}(g_{x}, g_{y})>1} N_{T}(g_{x}, g_{y}) *
log ( N_{T}(g_{x}, g_{y}) - 1 - b )\\
 & + & n_{1,T} * log ( \frac{b*(n_{+, T}-1)}{(n_{0,T}+1)} ) \nonumber \\
 & - & \sum_{g_{x}} N_{T}(g_{x}) * log (  N_{T}(g_{x}) - 1 ) - \sum_{g_{y}}
N_{T}(g_{y}) * log (  N_{T}(g_{y}) - 1 )
\nonumber .
\end{eqnarray}

\subsection{Clustering Algorithm}
\label{algo}

Given the $F'''_{LO}$ maximization criterion, we use the algorithm in Figure
\ref{fig:algo} to find a good clustering
function $G$. The algorithm tries to make local changes by moving words between
classes, but only if
it improves the value of the optimisation function. The algorithm will converge
because the
optimisation criterion is made up of logarithms of probabilities and thus has
an upper limit and
because the value of the optimisation criterion increases in each iteration.
However, the solution
found by this greedy algorithm is only locally optimal and it depends on the
starting conditions.
Furthermore, since the clustering of one word affects the future clustering of
other words, the
order in which words are moved is important. As suggested in \cite{Kne93}, we
sort the words by
the number of times they occur such that the most frequent words, about which
we know
the most,  are clustered first. Moreover, we do not consider infrequent words
(e.g. words
with occurrence counts smaller than $5$) for clustering, because the
information they
provide is not very reliable. Thus, if we start out with an initial clustering
in which no cluster
occurs only once, and if we never move words that occur only once,
then we will never have a cluster which occurs only once.
Thus, the assumption we made earlier,
when we decided to estimate cluster uni-grams by frequency counts, can be
guaranteed.
\begin{figure}
\begin{verbatim}
-start with an initial clustering function G
-iterate until some convergence criterion is met
{
     -for all x in  X and y in Y
     {
          -for all gx and gy
          {
               -calculate the difference in F(G) when
               x/y is moved from its current cluster to gx/gy
          }
          -move the x/y that results in the biggest improvement
          in the value of F(G)
     }
}
\end{verbatim}
\caption{The clustering algorithm}
\label{fig:algo}
\end{figure}

We will now determine the complexity of the algorithm. Let $M_{X}$ and $M_{Y}$
be
the maximal number of clusters for $X$ and $Y$, let $|X|$ and $|Y|$  be the
number
of possible values for $X$ and $Y$, let $M=max(M_{X}, M_{Y})$,
$W=max(|X|, |Y|)$ and let $I$ be the number of iterations.
When we move $x$ from $g_{x}$ to $g'_{x}$ in the inner loop
(the situation is symmetrical for $y$), we need
to change the counts $N(g_{x}, g_{y})$ and $N(g'_{x}, g_{y})$ for all $g_{y}$.
The amount by which we need to change the counts is equal to the
number of times $X$ occurred with cluster $g_{y}$. Since this amount is
independent of $g'_{x}$, we need to calculate it only once for each $x$.
The amount can then be looked up in constant time within the loop, thus
making the inner loop of order $M$. The inner loop is executed once for every
cluster $x$ can be moved to, thus giving a complexity of the order of $M^{2}$.
For
each $x$, we needed to calculate the number of times $x$ occurred with all
clusters $g_{y}$. For that  we have to sum up all the bigram counts
$N(x,y):G_{2}(y)=g_{y}$, which is on the order of $W$, thus giving a complexity
of the order
of  $W+M^{2}$. The two outer loops are executed $I$ and $W$ times thus giving
a total complexity of the order of $I*W*(W+M^{2})$.

\section{Results}
\label{results}

In the experiments performed so far, we work with $200,000$ verb-object pairs
extracted from
the 1989 section of the Wall Street Journal corpus. The data contains about
$19,000$
different direct object tokens, about $10,000$ different verb tokens and about
$140,000$ different
token pairs. We use $\frac{3}{4}$ of the data as training and the rest as
testing data. For
computational reasons, we have so far restricted the number of possible
clusters to 50, but
we hope to be able to increase that in the future. The perplexity on the
testing
text using the clustering algorithm on the verb-object pairs is shown in Table
\ref{tab:pp}.
For comparison, the table also contains the perplexity of a normal uni-gram
model
(e.g. no predictor variable $X$) and the performance of the clustering
algorithm on the
usual bi-gram data (e.g. the word immediately preceding the direct object as
predictor variable $X$).
We can see that the verb contains considerable information about the direct
object and
leads to a reduction in perplexity of about 18\%. However, the immediately
preceding word
leads to an even bigger reduction of about 34\%.
We also tried a linear interpolation of the two clustered models
\begin{equation}
p_{interpol}(y_{l})= \lambda * p_{verb-object}(y_{l}) + (1-
\lambda)*p_{bigram}(y_{k}).
\end{equation}
On a small set of unseen data,
we determined the best value (out of 50 possible values in $]0,1[$)
of the interpolation parameter $\lambda$.
As shown in Table \ref{tab:pp}, the interpolated model leads to an overall
perplexity reduction of 43\% compared to the uni-gram, which corresponds to a
reduction of about 10\%
over the normal bi-gram perplexity.
\begin{table}
\centering
\begin{tabular}{|c|c|c|} \hline
Model & Perplexity & \% Reduction\\ \hline
uni-gram & 3200 & not appl. \\
clustered verb-object & 2640 & 18\%\\
clustered bigram & 2010 & 37\%\\
combined model & 1820 & 43\%\\ \hline
\end{tabular}
\caption{Perplexity results}
\label{tab:pp}
\end{table}

\section{Conclusions}
\label{conclusions}

{}From a purely linguistic perspective, it would be slightly surprising to find
out that
the word immediately preceding a direct object can be used better to predict it
than the preceding verb. However, this conclusion can not  be drawn
from our results because of the noisy nature of the data. In other words, the
data contains pairs like (is, chairman), which would usually not
be considered as a verb-direct object pair. It is possible, that more accurate
data (e.g.
fewer, but only correct pairs) would lead to a different result. But the
problem with fewer pairs would of course be that the model can be used in fewer
cases, thus reducing the usefulness to a language model that
would predict the entire text (rather than just the direct objects). The
results
thus support the common language modeling practice, in that bi-gram events (by
themselves)
seem to be more useful than this linguistically derived predictor (by itself).
Nevertheless,
the interpolation results also show that this linguistically derived
predictor is useful as a complement to a standard class based bigram model.
In the future, we hope to consolidate these early findings
by more experiments involving a higher number of clusters and a larger data
set.

\end{document}